# AN ALGEBRAICAL MODEL FOR GRAY LEVEL IMAGES

**Vasile PATRASCU**
Department of Informatics Technology, Romanian Air Transport (TAROM)
e-mail: vpatrascu@tarom.ro
**Ioan VOICU**
Advanced Technologies Institute
e-mail: nvoicu@ita.gov.ro

**Abstract**: *In this paper we propose a new algebraical model for the gray level images. It can be used for digital image processing. The model adresses to those images which are generated in improper light conditions (very low or high level). The vector space structure is able to illustrate some features into the image using modified level of contrast and luminosity. Also, the defined structure could be used in image enhancement. The general approach is presented with experimental results to demonstrate image enhancement*

## 1. Introduction.

We present two classical solutions of the problem. The first belongs to Alan V. Oppenheim who proposed the first algebraical model in 1965.
Oppenheim analyses non-linear systems, specially homomorphic systems using superposition and generalized addition. He did not used algebraic operations in the gray level images. Instead of that, a new representation of real numbers was used and classical operations.
The main characteristics of this model are:
- Gray level interval $\quad E = (0, \infty)$
- Addition $\quad \forall u, v \in E, \quad u \langle + \rangle v = u \cdot v$
- Neutral element $\quad \forall u \in E, \quad u \langle + \rangle e = e \langle + \rangle v = u$
with e=1

- Subtraction $\quad \forall u, v \in E, \quad u \langle - \rangle v = \dfrac{u}{v}$

- Negative element $\quad \forall u \in E, \quad \langle - \rangle u = \dfrac{1}{u}$

- Scalar multiplication
$\forall \lambda \in R, \forall u \in E, \quad \lambda \langle \times \rangle u = u^{\lambda}$
- Isomorphism $\quad \phi : E \to R, \quad \phi(x) = \ln(x)$
and the inverse $\quad \phi^{-1} : R \to E, \quad \phi^{-1}(y) = e^{y}$

The Oppenheim's model realizes the homomorphic filtering.

The second model belongs to Michel Jourlin and Jean-Charles Pinoli. They proposed their algebraical model in 1985. This new algebraic and functional frame so called LIP (logarithmic image processing) allowed the introduction of addition between two images, interpolation and a sequence of images starting from an image to another, dynamic range enhancement. Their model defines abstract gray level range to be $(-\infty, M)$, real numbers when M>0. The above range contains the interval (0,M) which represents the concret gray levels used in image representation. The utility of negative values is purely theoretic. The subtraction operation is not always possible, so that the number of operations for logarithmic images is limited. The stability of addition operation does not verify the properties of the group structure defined on (0,M). Scalar multiplication is limited to positive values, being stable only on the interval $(0, \infty)$. Because a vector space structure can not be defined, the number of real aplications is reduced.

The main characteristics of this model are:
- Abstract gray level interval $\quad E = (-\infty, M)$
- Concrete gray level interval $\quad$ (0,M) $\quad$ with M>0

- Addition $\quad \forall u, v \in E, \quad u \langle + \rangle v = u + v - \dfrac{u \cdot v}{M}$

- Subtraction $\quad \forall u, v \in E, \quad u \langle - \rangle v = M \cdot \dfrac{u - v}{M - v}$

- Neutral element
$\quad \forall u \in E, \quad u \langle + \rangle e = e \langle + \rangle v = u \quad$ with e=0

- Negative element $\quad \forall u \in E, \quad \langle - \rangle u = - \dfrac{M \cdot u}{M - u}$

- Scalar multiplication

$\forall \lambda \in R, \forall u \in E, \lambda \langle \times \rangle u = M \cdot \left( 1 - \left( 1 - \dfrac{u}{M} \right)^{\lambda} \right)$

- Isomorphism $\varphi : R \to E, \quad \phi(x) = \ln \left( \dfrac{M}{M - x} \right)^{M}$

and the inverse

$$\phi^{-1}: E \to R, \quad \phi^{-1}(y) = M \cdot \left(1 - e^{-\frac{y}{M}}\right)$$

All this observations lead us to the proposal of a new algebraical model for gray level images.

The reminder of the paper is organized as follows: section 2 introduces our new algebraical model, section 3 presents some aspects of our experiments and section 4 summaries the results.

## 2. The vector space of the gray levels.

Let be M a positive real number. An image is represented by real, positive function which takes values in an interval [0,M]. For being able to deal with subtraction we will accept as a range of the image values the interval (-M,M). This function is defined on a spatial support $D \subset R^2$ and its values are from the interval $E = (-M, M)$ called the set of the gray levels. We usually choose M=1.

### 2.1. Addition.

We define the addition operation on the gray level interval by the following relation:

$$\forall u, v \in E, \quad u\langle+\rangle v = \frac{u+v}{1 + \frac{u \cdot v}{M^2}}$$

This operation has the following four properties:
(i) Stability $\quad \forall u, v \in E, \quad w = u\langle+\rangle v \in E$
(ii) Associativity
$\forall u, v, w \in E, \quad (u\langle+\rangle v)\langle+\rangle w = u\langle+\rangle(v\langle+\rangle w)$
(iii) Commutativity
$\quad \forall u, v \in E, \quad u\langle+\rangle v = v\langle+\rangle u$
(iv) Neutral element $\quad \forall u \in E, \quad u\langle+\rangle 0 = 0\langle+\rangle u$

### 2.2. Scalar multiplication.

Let us define an external multiplication of a gray level by the formula:
$\forall \lambda \in R, \forall u \in E,$

$$\lambda\langle\times\rangle u = M \cdot \frac{(M+u)^\lambda - (M-u)^\lambda}{(M+u)^\lambda + (M-u)^\lambda}$$

This operation has the following properties:
(i) $\quad \forall u \in E, \quad 0\langle\times\rangle u = 0$
(ii) $\quad \forall \lambda \in R, \quad \lambda\langle\times\rangle 0 = 0$

(iii) $\forall \lambda, \mu \in R, \forall u \in E,$
$(\lambda \cdot \mu)\langle\times\rangle u = \lambda\langle\times\rangle(\mu\langle\times\rangle u)$

(iv) $\forall \lambda, \mu \in R, \forall u \in E,$
$(\lambda + \mu)\langle\times\rangle u = (\lambda\langle\times\rangle u)\langle+\rangle(\mu\langle\times\rangle u)$

(v) $\forall \lambda \in R, \forall u, v \in E,$
$\lambda\langle\times\rangle(u\langle+\rangle v) = \lambda\langle\times\rangle u\langle+\rangle\lambda\langle\times\rangle v$

### 2.3. Subtraction.

The subtraction operation will be defined by the formula:

$$\forall u, v \in E, \quad u\langle-\rangle v = \frac{u-v}{1 - \frac{u \cdot v}{M^2}}$$

It verifies the property:
$\forall u \in E, \quad (-1)\langle\times\rangle u = -u = 0\langle-\rangle u = \langle-\rangle u$

After these definitions we can state the folowing theorem:

### 2.4 Theorem:

The set E of gray levels is a real vector space for the operations $\langle+\rangle$ and $\langle\times\rangle$.
The proof is obvious due to the properties listed above.

### 2.5. Asymptotical behavior of the algebraical operation $\langle+\rangle, \langle\times\rangle$.

In the neighborhood of zero value we can say that the algebraical operation $\langle+\rangle, \langle\times\rangle$ has an asymptotical behavior like the classical operation $+$, $\times$ for real numbers. These properties are true because of the following approximation,

$1 >> \frac{u \cdot v}{M^2}$ for the small gray levels.

### 2.6. The fundamental isomorphism.

We have the following isomorphism between the real vectorial space of gray level ( $E, \langle+\rangle, \langle\times\rangle$ ) and the space $(R, +, \times)$ of real numbers :

$$\phi: E \to R, \quad \phi(x) = \ln\left(\frac{M+x}{M-x}\right)^{\frac{M}{2}}$$

with the inverse

$$\phi^{-1}: R \to E, \quad \phi^{-1}(y) = M \cdot \frac{e^{\frac{y}{M}} - e^{-\frac{y}{M}}}{e^{\frac{y}{M}} + e^{-\frac{y}{M}}}$$

The isomorphism $\phi$ verifies the two next important properties:
(i) $\quad \forall u, v \in E, \quad \phi(u\langle+\rangle v) = \phi(u) + \phi(v)$
(ii) $\quad \forall \lambda \in R, \forall u \in E, \quad \phi(\lambda\langle\times\rangle v) = \lambda \cdot \phi(u)$

## 3. Experimental results

Three images are used to illustrate the proposed algebraical model. Fig. 1a presents the first original image. Fig. 1b represents the modified contrast by scalar multiplication and fig. 1c represents a new

The 7th International Conference, Exhibition on Optimization of Electrical and Electronic Equipment, OPTIM 2000, Braşov, România
11-12 May, 2000

image obtained by addition of a constant. We can see that luminosity is changed.

Fig. 2a represents the second original image. Fig. 2b and 2c represent modified luminosities using two addition operation.

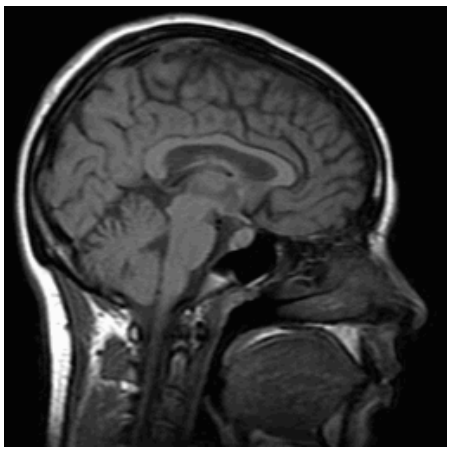

Fig.1a: Image of function $f_1$

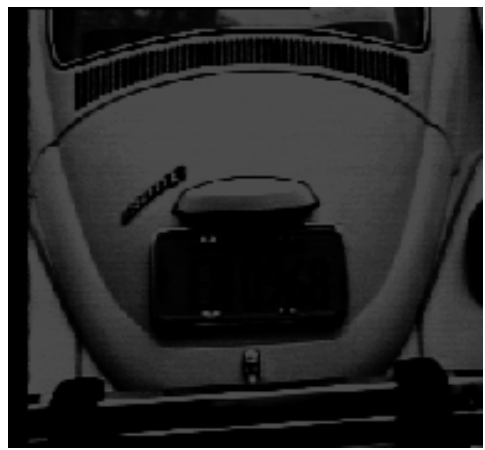

Fig. 2a: Image of function $f_2$

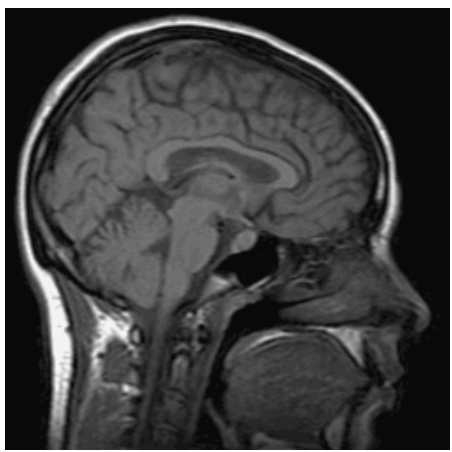

Fig.1b: Image of function $g = 0.8 \langle \times \rangle f_1$

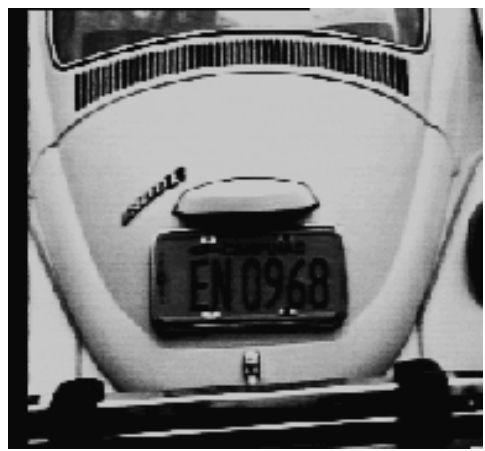

Fig. 2b: Image of function $g = f_2 \langle + \rangle 0.8$

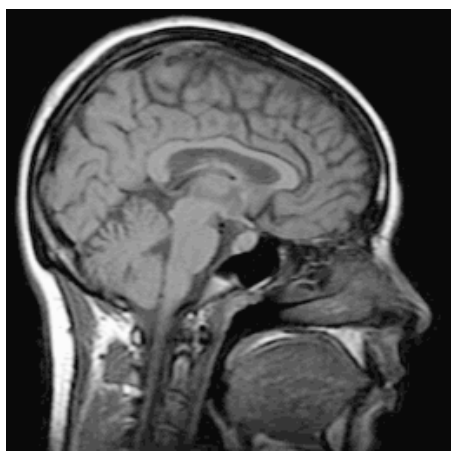

Fig. 1c: Image of function $g = f_1 \langle + \rangle 0.4$

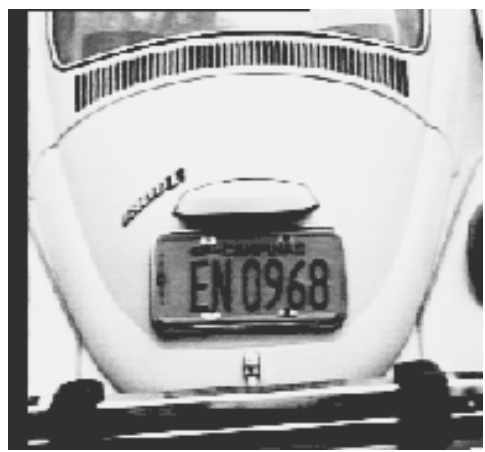

Fig. 2c: Image of function $g = f_2 \langle + \rangle 0.96$

Fig. 3a represents the third original image. Fig. 3b is the negative version of the image. Fig. 3c and 3d represent the images using the modul, respectively the signum function.

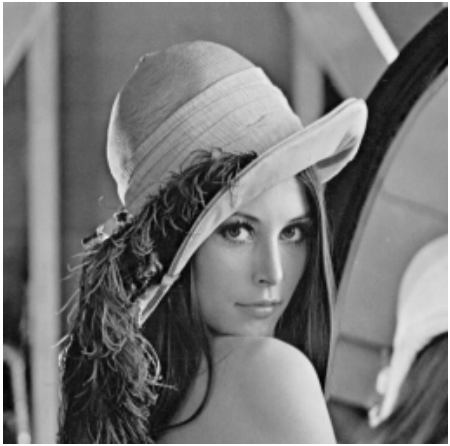

Fig. 3a: Image of function $f_3$

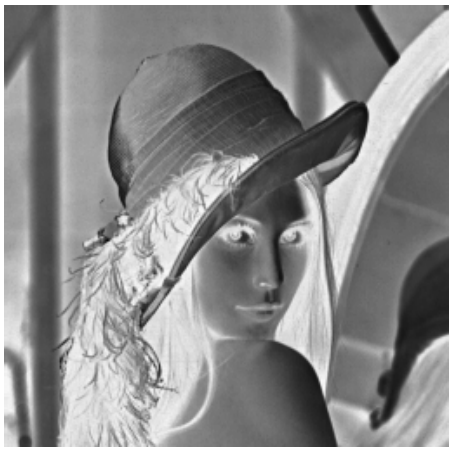

Fig.3b: Image of function $-f_3$

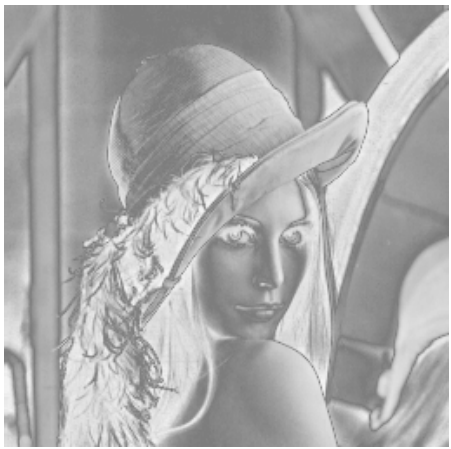

Fig. 3c: Image of function $|f_3|$

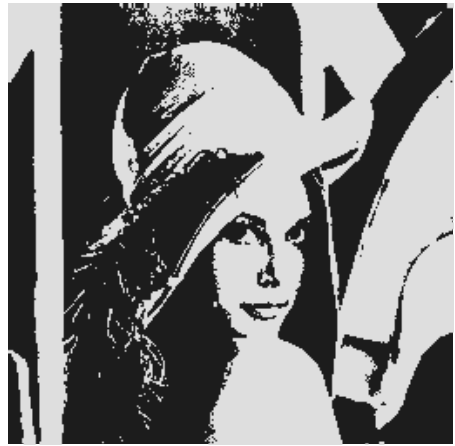

Fig. 3d: Image of function $\text{sign}(f_3)$

## 4. Conclusions.

We have presented in this paper a new algebraical model for gray level images. The main idea is to define an algebraical structure on a bounded interval. This structure is isomorph with real numbers and this isomorphism is a logarithmical one. The proposed model is very closely related to the human visual perception.